\documentclass{article}

\usepackage{microtype}
\usepackage{graphicx}
\usepackage{subcaption}
\usepackage{booktabs} 
\usepackage{makecell}
\usepackage{array}

\usepackage{placeins}

\usepackage{hyperref}

\usepackage[preprint]{icml2026}

\usepackage{amsmath}
\usepackage{amssymb}
\usepackage{mathtools}
\usepackage{amsthm}

\usepackage[capitalize,noabbrev]{cleveref}

\theoremstyle{plain}

\theoremstyle{definition}

\theoremstyle{remark}

\icmltitlerunning{Compositional Sparsity as an Inductive Bias for Neural Architecture Design}

\begin{document}

\twocolumn[
  \icmltitle{Compositional Sparsity as an Inductive Bias for Neural Architecture Design}




\begin{icmlauthorlist}
  \icmlauthor{Hongyu Lin}{ucl}
  \icmlauthor{Antonio Briola}{ucl}
  \icmlauthor{Yuanrong Wang}{ucl}
  \icmlauthor{Tomaso Aste}{ucl}
\end{icmlauthorlist}

\icmlaffiliation{ucl}{Department of Computer Science, University College London, London, United Kingdom}

\icmlcorrespondingauthor{Hongyu Lin}{hongyu.lin.18@ucl.ac.uk}


  \vskip 0.3in
]



\printAffiliationsAndNotice{}  

\begin{abstract}
\label{sec:Abstract}

Identifying the structural priors that enable Deep Neural Networks (DNNs) to overcome the curse of dimensionality is a fundamental challenge in machine learning theory. Existing literature suggests that effective high-dimensional learning is driven by compositional sparsity, where target functions decompose into constituents supported on low-dimensional variable subsets. To investigate this hypothesis, we combine Information Filtering Networks (IFNs), which extract sparse dependency structures via constrained information maximisation, with Homological Neural Networks (HNNs), which map the inferred topology into fixed-wiring sparse neural graphs. We formalise the design principles underlying this construction and present an interpretable pipeline in which abstraction emerges through hierarchical composition. HNNs are orders of magnitude sparser than standard DNNs and require only minimal hyperparameter tuning. On synthetic tasks with known sparse hierarchies, HNNs recover the underlying compositional structure and remain stable in regimes where dense alternatives degrade as dimensionality increases. Across a broad suite of real-world datasets, HNNs consistently match or outperform dense baselines while using far fewer parameters, exhibiting lower variance and showing reduced sensitivity to hyperparameters.

\end{abstract}

\section{Introduction}
\label{sec:Introduction}
Despite their remarkable empirical success, our understanding of Deep Neural Networks (DNNs) remains incomplete, particularly in regimes where classical learning theory predicts fundamental limitations \cite{zhang2017rethinking,razavi2021deep}. In the absence of structural assumptions on the target function, approximation and estimation errors for broad classes of functions on \(\mathbb{R}^p\) grow exponentially with the ambient dimension \(p\), reflecting the curse of dimensionality \citep{poggio2017and, danhofer2025position}.
A growing body of theoretical and empirical work has recently focused on identifying which forms of structure are implicitly exploited by modern architectures to escape this worst-case scenario \citep{beneventano2024neural, poggio2024compositional, danhofer2025position}. 
One recurring conclusion is that functions arising in real data admit sparse dependency patterns, where only small subsets of variables directly interact \citep{wang2023homological, briola2023homological,lin2024granger,aste2025information}. Moreover, these interactions are typically organised through hierarchical compositions across scales \citep{poggio2017and, poggio2025efficiently, danhofer2025position}.
From this point of view, the central issue in high-dimensional learning concerns the ability to capture, represent and exploit sparse, multi-scale interaction structures without incurring combinatorial complexity \citep{aste2025information, danhofer2025position}.

Recent theoretical work sharpens this view by arguing that compositional structure, together with symmetries and the reuse of intermediate representations, provides a concrete mechanism by which deep models can avoid worst-case scaling and generalise in regimes where unstructured alternatives fail \citep{vig2019analyzing, murty2022characterizing, petty2024impact, lindsey2025biology, danhofer2025position}. From this standpoint, the key question is not whether compositional sparsity is beneficial, but how to instantiate it as an inductive bias without relying on extensive architectural trial-and-error \citep{elsken2019neural, wang2023homological, briola2023homological, briola2024deep}. In real-world data, the relevant groups of interacting variables are unknown, and the space of possible higher-order interactions grows combinatorially with dimension \citep{aste2025information, aste2025probabilistic}. Standard deep learning workflows address this challenge implicitly through architecture and hyperparameter search over depth, width, and connectivity \citep{feurer2019hyperparameter, he2021automl}, sometimes followed by pruning and compression \citep{lecun1989optimal, hassibi1992second, hinton2015distilling, han2015learning, frankle2019lottery}. As a consequence, empirical performance can be highly sensitive to the tuning protocol and available computational budget, obscuring whether observed gains reflect appropriate inductive bias or extensive search over model configurations \citep{grinsztajn2022tree}.

\noindent\textbf{Our Approach}
This paper takes an alternative approach, moving from heuristic design toward architectures whose topology is \emph{deterministically induced} from estimated data dependencies. We employ \emph{Information Filtering Networks} (IFNs) \citep{mantegna1999hierarchical, aste2005complex, tumminello2005tool, barfuss2016parsimonious, massara2016network, massara2019learning, briola2022dependency, briola2023topological, wang2023topological} to extract sparse dependency structures by optimising an information retention objective under strict topological constraints, yielding hierarchies of overlapping higher-order interactions that enhance interpretability. Building on this structure, we leverage \emph{Homological Neural Networks} (HNNs) \cite{wang2023homological, briola2023homological, briola2024deep, briola2025hlob}, which are sparse feedforward models that map the inferred IFN topology into fixed-wiring sparse neural graphs. This construction enforces a hierarchical interaction structure, with abstraction increasing systematically with network's depth. At each layer, higher-order units aggregate signals from lower-order constituents that co-occur in the data-induced dependency topology. Because the wiring is inferred from observations, structural design and parameter learning are decoupled: the architecture is fixed by the dependency structure, while parameters are optimised end-to-end. Under matched parameter budgets, HNNs often achieve better empirical performance than dense multilayer perceptron (MLP) baselines on structured tasks. The resulting pipeline is interpretable, stable, and operates in a minimal-tuning regime, enabling a rigorous evaluation of whether structural priors, rather than extensive hyperparameter optimisation, drive performance. From an architectural perspective, HNNs are related to broader work on sparse neural networks, where restricting connectivity serves as an inductive bias to improve generalisation and scalability \citep{mostafa2019dynamic, frankle2019lottery, evci2020rigl, ebli2020simplicial}.

\noindent\textbf{Empirical Validation}
We validate this approach in two complementary settings, treating HNNs as a data-driven construction, rather than an arbitrary architectural choice. First, on synthetic problems with known sparse hierarchies, we demonstrate that when the true interaction topology is provided via an oracle, the corresponding HNN consistently outperforms all baselines, confirming that the architecture can effectively exploit compositional sparsity when available. We then show that IFN-estimated structures, while imperfect, capture sufficient dependency information to maintain substantial performance advantages over dense architectures, particularly as dimensionality increases. Second, across a broad suite of real-world tabular regression datasets where ground-truth structure is unknown, we assess whether data-driven HNNs yield consistent gains under shared hyperparameters. Across both settings, the resulting architectures remain stable in regimes where dense DNN baselines degrade, achieving competitive or superior performance with far fewer trainable parameters, lower variance, reduced sensitivity to hyperparameters, and slow degradation in high-dimensional, low-sample regimes.

\noindent\textbf{Paper Organisation}
Section~\ref{s.prelim} introduces the background on compositional structure and IFNs, and reviews the construction of \emph{Maximally Filtered Clique Forests}.
Section~\ref{s.hnn} presents the methodology, describing how data-driven dependency structures are translated into sparse neural architectures and how Homological Neural Networks are constructed and trained.
Section~\ref{exp_synthetic} evaluates the approach on synthetic data with known interaction structure, allowing controlled analysis of scalability and robustness in high-dimensional regimes.
Section~\ref{exp_ctr23} reports results on the OpenML-CTR23 \cite{fischer2023openml} benchmark, assessing performance on real-world tabular regression tasks under a minimal-tuning protocol.
Finally, Section~\ref{conclusion} concludes with a discussion of the main findings and directions for future work.

\section{Background and Preliminaries}
\label{s.prelim}

\subsection{Compositional Structure and Approximation}



We adopt the standard assumption that high-dimensional targets admit sparse,
compositional structure, in which only small subsets of variables interact
directly and higher-order effects are formed by hierarchical composition
\cite{poggio2017and}.

Let's consider functions of the form
\begin{equation}
    f(x_1,\dots,x_p)
    =
    h\!\left(
    g_1(\mathbf{x}_{U_1}),\ldots,g_m(\mathbf{x}_{U_m})
    \right),
    \label{eq:compositional-form}
\end{equation}
where each $U_i \subset \{1,\dots,p\}$ indexes a (typically small) subset of
variables, $\mathbf{x}_{U_i} \in \mathbb{R}^{|U_i|}$ denotes the corresponding
sub-vector, each $g_i : \mathbb{R}^{|U_i|} \to \mathbb{R}$ models an 
interaction, and $h : \mathbb{R}^m \to \mathbb{R}$ combines these components.
Under standard smoothness assumptions on $g_i$ and $h$, approximation results show that achieving a given error depends on the maximal interaction order $\max_i |U_i|$, rather than the ambient dimension $p$, thereby mitigating the curse of dimensionality \cite{mhaskar2016deep,poggio2017and,poggio2024compositional}.

In this work, we focus on data-driven mechanisms for estimating such interaction structure and incorporating it into neural architectures.

\subsection{Information Filtering Networks}
\label{s.ifn}

Information Filtering Networks (IFNs) provide a framework for extracting sparse dependency structures from multivariate and high-dimensional data \cite{tumminello2005tool,aste2025information}.
IFNs build a dependency graph $\mathcal{G}=(V,E)$, where $V=\{1,\dots,p\}$ indexes the variables and $E \subset V\times V$ denotes the set of edges, by retaining the most informative relationships while satisfying explicit topological constraints such as tree \cite{chow1968trees} or chordal graph \cite{blair1993introduction}.
These constraints act as a form of structural regularisation \cite{srebro2003bounded}, controlling model complexity and improving robustness in high-dimensional settings. A key feature of IFNs is that sparsification is achieved through a global optimisation criterion rather than through penalizing dense structures such as through $\ell_1$-regularised likelihood estimation \cite{banerjee2008model}. As a result, the retained structure concentrates the strongest dependencies compatible with the chosen constraints, yielding networks that are globally sparse but locally dense, with an interpretable and stable structure \cite{aste2025information}.

\paragraph{Maximally Filtered Clique Forest (MFCF)}
The Maximally Filtered Clique Forest (MFCF) \cite{massara2019learning} is a type of Markov Random Field \cite{kindermann1980markov,koller2009probabilistic} within the IFN family, obtained by restricting the admissible models to decomposable chordal graphs \cite{blair1993introduction, massara2019learning}. MFCF represents higher-order dependencies through a set of overlapping maximal cliques linked by separators, forming a clique forest. 

Starting from a single node, at each iteration, MFCF considers expanding the current decomposable graph by introducing a new maximal clique \cite{lauritzen1996graphical,massara2019learning}. Each candidate update is characterised by a pair $(C,S)$, where $C$ denotes the proposed clique formed by adding a new variable to an existing separator and $S \subset C$. The separator must be contained in at least one current clique, ensuring that the resulting graph remains decomposable.
The quality of such an update is evaluated using the local gain function
\begin{equation}
    \Delta(C,S)
    \;=\;
    \mathcal{I}(X_C) - \mathcal{I}(X_S),
    \label{eq:mfcf-gain}
\end{equation}
where $\mathcal{I}(\cdot)$ is an information measure computed from the data.
Only expansions with positive gain are accepted, which ensures that each added clique captures additional statistical dependence not already captured by its separator $S$. 

To control graph complexity and interpretability, MFCF relies on a minimum of two user-defined parameters. A \textit{gain threshold} determines whether a candidate expansion is accepted. A \textit{maximal clique size} ($K$) can be imposed to limit the highest interaction order retained by MFCF for sparsity control. 


MFCF provides a data-driven approximation to the full dependency structure by retaining the most informative higher-order interactions compatible with the model class. Although MFCF is not the only approach to structure estimation, it complements widely used methods such as neighborhood selection \cite{meinshausen2006neighborhood}, graphical lasso \cite{friedman2008glasso}, and constrained precision matrix estimators \cite{cai2011clime}. In contrast to these approaches, MFCF emphasises topological sparsification and higher-order interactions rather than edge regularisation.

\section{Methodology}
\label{s.hnn}

We now describe how a dependency structure inferred from data is translated into a sparse neural architecture. 

\subsection{Dependency Matrix and Dependency Graph Construction}
\label{mfcf}
In this paper, we use IFNs, generated via MFCF construction, from a pairwise dependency measure
$M \in \mathbb{R}^{p \times p}$, where $M_{ij}$ quantifies the statistical dependence between variables $X_i$ and $X_j$.
Depending on the task, $M$ may be computed using standard dependency measures such as correlations, or using label-aware variants that condition on the training targets \cite{guyon2003introduction,koller2009probabilistic}.
All dependency estimates are computed exclusively on training data and remain fixed throughout training, avoiding information leakage \cite{hastie2009elements}.

The dependency matrix $M$ serves as input to MFCF.
In all experiments, we instantiate MFCF using squared linear correlations $M_{ij}=\mathrm{corr}(X_i,X_j)^2$ computed on the training fold, unless otherwise specified by a label-aware variant.
More general dependency measures, such as Mutual Information \cite{cover2006elements}, could equally be used within the same framework.
Here we adopt linear correlations to keep the structural inference step simple, and to avoid introducing additional modelling assumptions or hyperparameters. 

Applying MFCF to the dependency matrix produces a clique forest that captures the dominant dependency structure in the data. Throughout this work, we fix the maximal clique size to $K=4$ \cite{massara2016network}. 
This choice is not tuned for performance, but reflects a structural modelling assumption that limits interaction order to control combinatorial growth and computational cost. The resulting clique forest defines the connectivity structure of our neural architecture.




\subsection{From Clique Forests to Network Layers}
\label{s:hnn:simplicial}

\begin{figure*}[t]
  \centering
  \includegraphics[width=\textwidth]{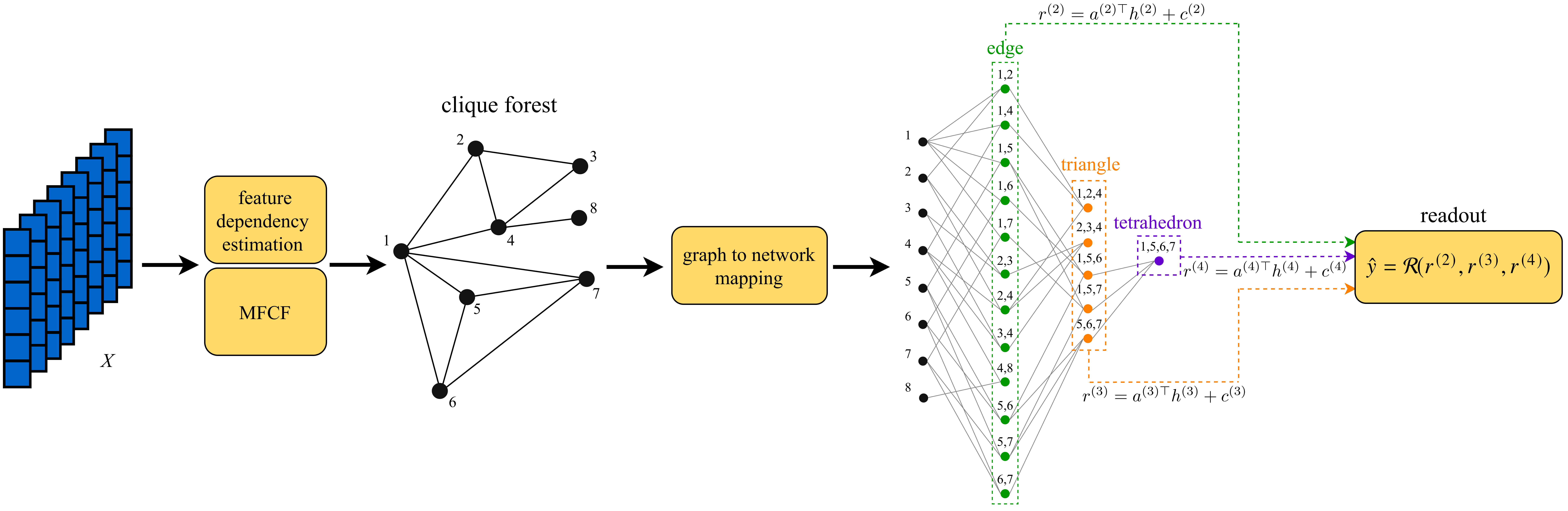}
  \vspace{-0.4em}
  \caption{
        Pipeline for constructing a HNN from data.
        Input features $X$ are first used to estimate a dependency matrix and construct a MFCF, yielding a decomposable clique tree.
        The clique structure induces a hierarchy of cliques, which is lifted into a sparse feed-forward neural architecture where units correspond to cliques of increasing order and connections follow incidence.
        A readout unit aggregates the resulting representations to produce the final prediction.
        }
  \label{fig:hnn-schema}
  \vspace{-0.6em}
\end{figure*}



\paragraph{IFN Higher-order Interactions}
The MFCF returns a collection of maximal cliques $\mathcal{C}$, each representing a set of variables involved in a data-supported dependency.
From these cliques, we construct a hierarchy of higher-order interaction units by including, for each clique $C \in \mathcal{C}$, all $k$-way variable subsets with $k \le K$.
Each such subset $\alpha \subseteq C$ defines a $k$-way interaction among the variables in $\alpha$.

This construction ensures that whenever a $k$-way interaction is included, all of its lower-order interactions are included as well.
Such hierarchical organisation reflects a standard assumption in compositional modelling, namely that higher-order interactions are built from their lower-order constituents \cite{poggio2017and}.

We denote by $\mathcal{H}^{(k)}$ the set of unique $k$-way interaction units induced
by the maximal cliques, i.e., all distinct subsets $\alpha$ with $|\alpha|=k$ and
$\alpha \subseteq C$ for some $C\in\mathcal{C}$.
The full collection of interaction units used to construct the network is
$\mathcal{H}=\bigcup_{k=1}^{K}\mathcal{H}^{(k)}$.
If the same subset $\alpha$ appears in multiple maximal cliques, it corresponds to
a single unit in the network.
\paragraph{Interaction Layers and Inclusion-based Wiring}
As shown in Figure~\ref{fig:hnn-schema}, for each $k\in\{1,\dots,K\}$, we instantiate one layer with one unit per $\alpha\in\mathcal{H}^{(k)}$.
A unit $\alpha\in\mathcal{H}^{(k)}$ receives inputs only from units in the preceding layer that correspond to subsets of $\alpha$. These constraints impose a strong inductive bias toward sparse, compositional interactions: variables are allowed to interact only when they are in the same clique supported by the inferred dependency structure.
As a consequence, the resulting networks are sparse by construction, with the number of trainable parameters 
scaling with the number of incidence relations rather than quadratically in the ambient dimension.
Moreover, the architecture is fully specified prior to training: the MFCF fully determines the set of maximal cliques and all connectivity masks. Architectural design is therefore decoupled from parameter optimisation, and training proceeds by learning only the nonzero weights end-to-end using the standard gradient-based methods \cite{rumelhart1986learning,kingma2015adam}.



\subsection{Layer-wise Computation and Parameterisation}

Let $h^{(k)} \in \mathbb{R}^{|\mathcal{H}^{(k)}|}$ denote the vector of activations at interaction order $k$. The network is evaluated in a feed-forward manner \cite{rumelhart1986learning,hornik1989multilayer} from lower to higher orders.
For $k \geq 2$, activations are computed as
\begin{equation}
    h^{(k)} = \sigma\!\left( W^{(k)} h^{(k-1)} + b^{(k)} \right),
\end{equation}
where $\sigma(\cdot)$ is a fixed element-wise nonlinearity \cite{goodfellow2016deep}.
The bias vector $b^{(k)} \in \mathbb{R}^{|\mathcal{H}^{(k)}|}$ is unconstrained. The weight matrix $W^{(k)} \in \mathbb{R}^{|\mathcal{H}^{(k)}| \times |\mathcal{H}^{(k-1)}|}$ is sparsely parameterised according to the interaction hierarchy: an entry $W^{(k)}_{ij}$ is trainable if and only if the order-$(k{-}1)$ interaction indexed by $j$ is a subset of the $k$-way interaction indexed by $i$. All other entries are fixed to zero and excluded from optimisation.

\subsection{All-Layer Readout Mechanism}

Unlike conventional feed-forward architectures which typically form predictions from
the last hidden layer~\cite{goodfellow2016deep}, HNNs aggregate information from all
interaction layers. This design is inspired by layer-wise aggregation mechanisms
explored in other deep learning settings, where intermediate representations are
combined to mitigate information loss across depth
\cite{lee2015deeply,xu2018jumping,hamilton2020graph}.

For each interaction order $k \geq 2$, we attach a readout that maps the corresponding
activations to a scalar. In this work, we adopt a simple linear form,
\begin{equation}
r^{(k)} = a^{(k)\top} h^{(k)} + c^{(k)},
\end{equation}
where $a^{(k)} \in \mathbb{R}^{|\mathcal{H}^{(k)}|}$ and $c^{(k)} \in \mathbb{R}$ are
trainable parameters.

The scalars $\{r^{(k)}\}_{k=2}^{K}$ are then combined through a readout operator $\mathcal{R}$ to produce prediction:
\begin{equation}
\hat{y} = \mathcal{R}\!\left(r^{(2)}, \ldots, r^{(K)}\right).
\end{equation}
In this work, we choose a linear instantiation,
\begin{equation}
\hat{y} = \sum_{k=2}^{K} \beta_k\, r^{(k)} + d,
\end{equation}
with learned coefficients $\{\beta_k\}$ and bias $d$.

This all-layer readout complements the feed-forward interaction hierarchy by providing a direct aggregation path from each interaction order to the output.
While information propagates upward from lower to higher-order interactions through incidence-based connections, the readout explicitly preserves contributions from all interaction orders. This is particularly important when gain thresholding in MFCF suppresses weak dependencies, as some lower-order interactions may not participate in any higher-order interactions.
In such cases, lower-order representations are not propagated to deeper layers and would have no direct influence on the output.
We adopt a linear readout by design, as it preserves an additive decomposition across interaction orders and maintains a clear correspondence between interaction order and predictive contribution. More expressive nonlinear readouts are possible, but may entangle interaction orders and obscure this compositional structure.




\section{Experiments}
\label{exp}
We evaluate HNNs on both synthetic data and well-known tabular regression benchmarks.
Synthetic experiments allow to assess whether models can exploit known interaction structures under controlled conditions, while real-world benchmarks test whether the same inductive biases remain effective when the underlying structure is unknown.
In Section~\ref{exp_synthetic}, we present both quantitative and qualitative analyses comparing HNNs with baseline models across multiple dimensional settings with known data-generating processes.
In Section~\ref{exp_ctr23}, we evaluate HNNs on the OpenML-CTR23 tabular regression benchmark suite \cite{fischer2023openml} to assess robustness and predictive performance relative to the baselines.
\paragraph{HNN Variants}
Across all experiments, we study a family of HNNs that
share the same architectural template and training procedure, and differ in how the dependency matrix used by MFCF is constructed, which in turn determines the resulting
interaction hierarchy and parameter count.
We consider two data-driven instantiations:
\begin{itemize}
    \item \textsc{HNN (marginal).}
    The dependency matrix is constructed using squared Pearson correlations computed on the
    training fold, followed by MFCF with maximal clique size $K=4$.
    This is the simplest and most standard instantiation of the pipeline, serving as the
    primary structural baseline.

    \item \textsc{HNN (m-s).}
    This median-split variant is label-aware in the sense that training targets are
    used to stratify the data.
    We split the training set at the empirical median of the target, compute squared Pearson
    correlations separately on each split, apply MFCF ($K=4$) independently, and construct
    a single HNN from the union of the resulting interaction hierarchies.
    Since the union retains interactions discovered in both splits, this variant typically
    yields a larger interaction hierarchy and a larger parameter count than the marginal
    construction, and is designed to capture regime-dependent interactions under potential
    distribution shift~\cite{quinonero2009datasetshift,zliobaite2010drift,gama2014drift}.
\end{itemize}
\paragraph{Synthetic-only Structure-controlled Baselines}
In addition to the two data-driven HNN variants, the synthetic setting enables models that isolate the role of correct wiring from sparsity and model capacity:
\begin{itemize}
    \item \textsc{HNN (oracle).}
    The HNN structure is fixed to the ground-truth interactions used to generate the target (Appendix~\ref{app:synthetic}), and is therefore not constrained by the MFCF maximal clique size $K=4$.

    \item \textsc{HNN (rand oracle).}
    Connectivity is randomised while preserving the oracle layer-wise sparsity pattern and parameter budget, isolating the effect of correct structural alignment from sparsity alone.

    \item \textsc{MLP-HNN}.
    A dense baseline that preserves the hierarchical depth and readout structure of an HNN while replacing incidence-based sparse connectivity with fully connected layers.

    \item \textsc{MLP (PM-HNN)} and \textsc{MLP (PM-oracle).}
    Single-layer MLP baselines with trainable parameter counts matched to \textsc{HNN (marginal)} and \textsc{HNN (oracle)}, respectively.
\end{itemize}
These baselines are designed as controlled diagnostics. Together, they disentangle the effects of (i) correct interaction alignment (\textsc{HNN (oracle)} vs.\textsc{HNN (rand oracle)}), (ii) sparsity versus dense
(\textsc{HNN} vs.\ \textsc{MLP-HNN}), and (iii) architectural structure versus raw capacity
(parameter-matched MLPs), enabling a clean interpretation of performance differences in
the synthetic setting.

\paragraph{Baselines}
We compare HNNs against standard dense multilayer perceptrons (MLPs)~\cite{goodfellow2016deep} and tree ensembles, namely Random Forests~\cite{breiman2001random} and XGBoost~\cite{chen2016xgboost}.
We report both fixed-architecture MLPs and (when applicable) parameter-matched MLPs to control for model capacity.

\paragraph{Training Protocol}
Unless otherwise stated, all neural models are trained using Adam~\cite{kingma2015adam},
with ReLU activations~\cite{nair2010rectified}.
HNNs additionally employ layer normalisation~\cite{ba2016layer} as part of their
architectural design.
No dataset-specific or variant-specific hyperparameter tuning is performed.
A full description of optimisation settings and architectural hyperparameters is
provided in Appendix~\ref{app:hparams}. 

Predictive performance is evaluated using rankings induced by test $R^2$, within each
test fold, ranking models by $R^2$ is equivalent to ranking by MSE, since the total
target variance is fixed.
We aggregate ranks across folds and datasets to compare heterogeneous regimes in which
absolute error scales are not directly comparable.
Throughout, the MFCF maximum clique size is fixed to $K=4$ and treated as a capacity
control parameter, in order to isolate the effect of data-driven wiring.

\subsection{Synthetic Data}
\label{exp_synthetic}

Our synthetic experiments are based on a data generation process with known
interaction structure. Feature vectors are sampled from a sparse Gaussian graphical model
\cite{lauritzen1996graphical}, where conditional independencies are specified by an
Erd\H{o}s-R\'enyi random graph \cite{erdos1959random,janson2011random}, inducing
realistic correlations without explicitly fixing higher-order dependencies.
Targets are generated as sums of nonlinear functions acting on variable subsets
supported by cliques of the underlying dependency graph.
Whenever a higher-order interaction is present, all of its lower-order subsets
are also included, yielding targets with explicit hierarchical compositional
structure and heterogeneous interaction orders.
This construction allows us to study the effect of architectural inductive bias
under controlled sparsity and noise while retaining realistic dependencies
between inputs.
Full details of the data generation procedure are provided in
Appendix~\ref{app:synthetic}. Importantly, the maximal clique size \(K\) constrains only the maximal clique size admitted by MFCF and the resulting network architecture, without restricting the interaction orders used to generate the synthetic targets.

We vary both the feature dimension \(p\) and the sample size \(n\), spanning regimes from well-specified to highly underdetermined \((p/n \in [0.01, 3])\).
Because MFCF relies on pairwise dependence estimates, we highlight a data-rich regime ($p/n \le 0.05$) in which dependence estimation is reliable across the full range of feature dimensions considered ($p\in[50,5000]$). We report results in this regime in the main text; the full sweep over $p/n\in[0.01,3]$ is provided in Appendix~\ref{app:synthetic-full}.

\subsubsection{Results and Interpretation}
\begin{figure*}[t]
  \centering
  \includegraphics[width=\textwidth]{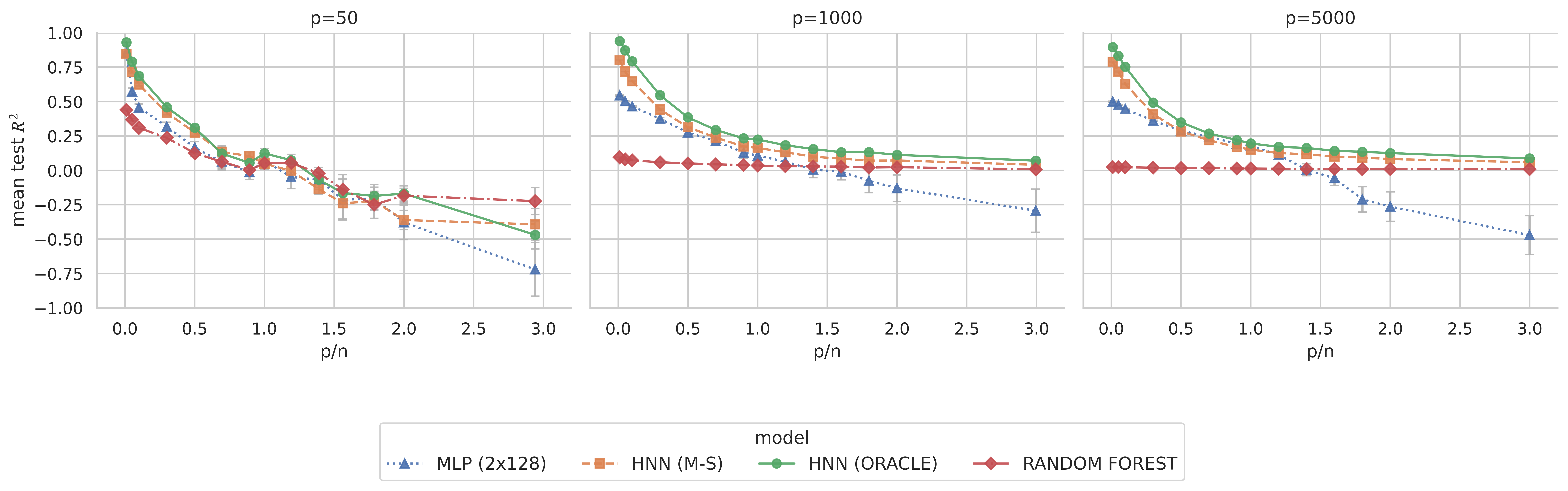}
  \vspace{-0.4em}
  \caption{
    Mean test $R^2$ as a function of the dimension-to-sample ratio $p/n$ for three representative feature dimensions ($p\in\{50,1000,5000\}$).
    For readability, we plot a subset of baselines here; full results across all methods are reported in Appendix~\ref{app:synthetic-full}. Each curve shows the mean test $R^2$ across runs at fixed $p/n$, with error bars indicating $\pm$SEM.
  }
  \label{fig:pn-focused}
\end{figure*}
\begin{table}[h]
  \caption{Test $R^2$ performance ranking on synthetic data in the data-rich regime ($p/n \le 0.05$) for $p\in[50,5000]$.
  For each $(p,n)$ configuration, results are averaged over 30 independent runs and ranks are then aggregated across configurations.
  Lower rank indicates better performance. Parameter ranges report the number of trainable parameters across $p$.}
  \label{tab:r2-ranks-rich}
  \centering
  \small
  \setlength{\tabcolsep}{3pt}
  \begin{sc}
  \begin{tabular}{lcc}
    \toprule
    Model & Mean rank ($\downarrow$) & Params \\
    \midrule
    HNN (oracle) & 1.00 & 1.3K--49.7K \\
    HNN (m-s) & 2.08 & 3.8K--144K \\
    HNN (marginal) & 3.33 & 2.2K--93.9K \\
    HNN (rand oracle) & 4.41 & 1.4K--49.7K \\
    MLP-HNN & 5.33 & 35.0K--59.9M \\
    MLP (2$\times$128) & 5.50 & 79.1K--578K \\
    MLP (PM-HNN) & 6.67 & 2.2K--94.1K \\
    MLP (PM-oracle) & 7.67 & 1.3K--50.1K \\
    XGBoost & 9.00 & -- \\
    Random Forest & 10.00 & -- \\
    \bottomrule
  \end{tabular}
  \end{sc}
  \vspace{-0.1in}
\end{table}



We begin by focusing on the data-rich regime ($p/n \le 0.05$), where the interaction structure inferred from pairwise dependencies is reliable across the considered feature dimensions.
Table~\ref{tab:r2-ranks-rich} reports aggregated test $R^2$ rankings for $p\in[50,5000]$. \textsc{HNN (m-s)} is the strongest data-driven variant (mean rank $2.08$), consistently outperforming \textsc{HNN (marginal)} and all dense MLP baselines.
Parameter-matched MLPs (\textsc{MLP (PM-HNN)} and \textsc{MLP (PM-oracle)}) remain substantially worse than their corresponding HNN variants, indicating that the performance gains cannot be explained by parameter count alone, but instead arise from the interaction-structured wiring imposed by the inferred hierarchy.

Structure-controlled baselines further clarify the role of data-aligned connectivity.
The randomised oracle (\textsc{HNN (rand oracle)}) preserves the oracle’s sparsity pattern and parameter budget but performs significantly worse, showing that sparsity alone is insufficient without correct wiring.
Conversely, \textsc{MLP-HNN} removes sparsity, while preserving hierarchical depth and readout, resulting in a much larger parameter count yet worse performance, showing that dense capacity does not recover the benefits of structured connectivity. Beyond this data-rich regime, we evaluate all methods across a broader range of dimension-to-sample ratios, including highly underdetermined settings ($p/n\in[0.01,3]$). The full set of results over this range is reported in Appendix~\ref{app:synthetic-full}.

Figure~\ref{fig:pn-focused} illustrates how test $R^2$ evolves with $p/n$ for representative feature dimensions ($p\in\{50,1000,5000\}$).
Across all three dimensions, dense MLP baselines exhibit a rapid degradation as $p/n$ increases, with performance collapsing sharply in low-sample regimes.
This effect becomes more pronounced as $p$ grows, highlighting the sensitivity of unstructured architectures to dimensionality when data are scarce \cite{wainwright2019highdimensional}.
Tree-based methods show greater robustness than dense MLPs but quickly plateau at near-zero test $R^2$, failing to benefit from additional samples in higher-dimensional settings.

In contrast, the \textsc{HNN (m-s)} degrades substantially more slowly as $p/n$ increases and maintains a clear separation from both dense MLP and tree-based baselines across all dimensions.
The \textsc{HNN (oracle)} forms an upper envelope throughout, confirming that when the correct interaction structure is known, the hierarchical architecture can exploit it uniformly across regimes.
The relatively small and stable gap between the \textsc{HNN (oracle)} and \textsc{HNN (m-s)} curves suggests that the data-driven construction captures a substantial fraction of the useful interaction structure, particularly in high-dimensional, low-sample settings.

\begin{figure}[h]
  \centering
  \includegraphics[width=\columnwidth]{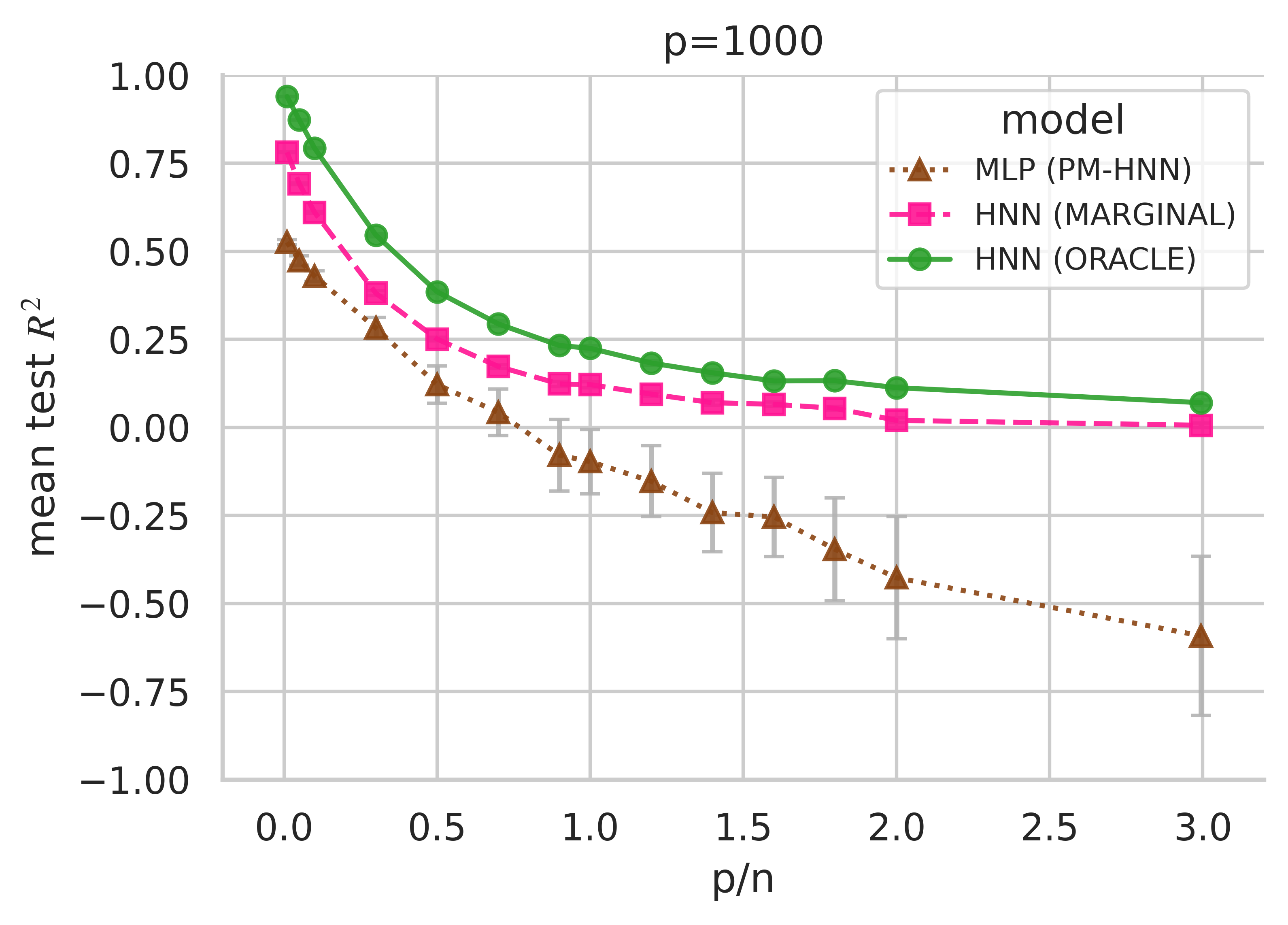}
  \vspace{-0.4em}
  \caption{
    Mean test $R^2$ as a function of the dimension-to-sample ratio $p/n$ for $p=1000$.
    We compare \textsc{HNN (marginal)} with a single-layer dense MLP whose number of trainable parameters is matched to that of the \textsc{HNN (marginal)}, together with the \textsc{HNN (oracle)}.
    Results are averaged over tasks and random seeds, with error bars indicating $\pm$SEM.
    }

  \label{fig:pn-1000}
\end{figure}

To further disentangle the effect of architectural structure from model capacity, Figure~\ref{fig:pn-1000} focuses on a representative high-dimensional setting with $p=1000$ and compares a \textsc{HNN (marginal)} to a single-layer dense MLP with a matched number of trainable parameters.
Despite having comparable capacity, the HNN consistently attains higher test $R^2$ and exhibits substantially greater stability as $p/n$ increases.
This confirms that the observed performance gains cannot be attributed to parameter count alone, but instead arise from the data-aligned sparsity and compositional structure imposed by the inferred architecture \cite{poggio2017and,belkin2019reconciling}.


\subsection{OpenML-CTR23 Tabular Regression Benchmark}
\label{exp_ctr23}

We complement the controlled synthetic experiments with evaluations on the OpenML-CTR23 \cite{fischer2023openml} curated tabular regression benchmark.
OpenML-CTR23 is a suite of real-world regression tasks designed for systematic comparison under standardised splits and evaluation, following similar design principles to widely used OpenML benchmark suites \cite{tschalzev2024data}.
It contains a diverse set of datasets spanning a broad range of sample sizes and feature dimensions, and has been adopted in recent work on tabular regression and foundation models \cite{holzmuller2024better,ma2025tabdpt,wu2025zero}.
In contrast to the synthetic setting, the underlying interaction structure is unknown and can vary substantially across datasets.

We use the official OpenML cross-validation split provided with each task \cite{vanschoren2014openml,fischer2023openml}.
For each fold, the training portion is further split into a training and validation set by randomly holding out 20\% of the training data, while the test fold is kept strictly untouched.
All preprocessing steps (e.g., scaling and normalisation) are fitted exclusively on the training subset and then applied unchanged to the validation and test data, in accordance with standard best practices to prevent information leakage \cite{hastie2009elements,kaufman2012leakage}.
This protocol ensures reproducibility and a fair comparison across models.


\paragraph{Minimal Tuning Regime}
All neural models are trained using a fixed set of shared hyperparameters across datasets, with no dataset-specific architecture search. The choice of this experimental protocol is intentional: the goal is to assess whether data-driven structural priors alone can yield competitive performance under a constrained tuning budget, rather than maximising benchmark scores through extensive hyperparameter optimisation.
Exact optimisation settings and architectural hyperparameters are reported in Appendix~\ref{app:hparams}.

\subsubsection{Results and Interpretation}

Table~\ref{tab:ctr23-r2-ranks} reports aggregated test $R^2$ ranks on the OpenML-CTR23 benchmark.
Under the shared minimal-tuning protocol, the \textsc{HNN (m-s)} achieves a mean rank of $3.77$, outperforming all MLP baselines and ranking competitively with tree-based methods. In particular, it substantially improves over the strongest dense MLP (2$\times$256), which attains a mean rank of $4.01$, as well as over the single-hidden-layer MLP (1$\times$512) with a rank of $4.54$. The \textsc{HNN (marginal)} performs slightly worse, with a mean rank of $4.37$. Model ranking for each dataset can be found in Appendix \ref{ctr23full}.

\begin{table}[h]
  \caption{OpenML-CTR23 test $R^2$ performance ranking across datasets.
  For each dataset, models are ranked by mean test $R^2$ over the provided folds and repeats;
  ranks are then averaged across datasets.
  Lower rank indicates better performance.}
  \label{tab:ctr23-r2-ranks}
  \centering
  \small
  \setlength{\tabcolsep}{3pt}
  \begin{sc}
  \begin{tabular}{lcc}
    \toprule
    Model & Mean rank ($\downarrow$) & Params (range) \\
    \midrule
    XGBoost & 3.14 & -- \\
    HNN (m-s) & 3.77 & 0.1k--13.9k \\
    Random Forest & 3.91 & -- \\
    MLP (2$\times$256) & 4.01 & 67.6k--114.9k \\
    MLP (2$\times$128) & 4.31 & 17.4k--41.1k \\
    HNN (marginal) & 4.37 & 0.1k--8.6k \\
    MLP (1$\times$512) & 4.54 & 3.6k--98.3k \\
    \bottomrule
  \end{tabular}
  \end{sc}
  \vspace{-0.1in}
\end{table}





The comparison with dense MLPs is particularly informative.
Despite using one-to-two orders of magnitude more trainable parameters than HNNs, MLPs consistently obtain worse average ranks.
This indicates that, in the limited-tuning regime considered here, architectural inductive bias plays a more decisive role than raw model capacity~\cite{poggio2024compositional}, consistent with prior findings that unstructured neural networks often underperform on tabular data without extensive tuning or architectural specialisation~\cite{grinsztajn2022tree,shwartz2022tabular}.

As expected on OpenML-CTR23, tree-based ensemble methods remain strong baselines~\cite{fischer2023openml,wu2025zero}.
XGBoost achieves the best overall average rank ($3.14$), followed closely by Random Forests ($3.91$).
Notably, the gap between XGBoost and the \textsc{HNN (m-s)} is small relative to the separation between either of these methods and standard MLPs, indicating that HNNs substantially narrow the performance gap between neural and tree-based models under matched evaluation conditions.
Among HNN variants, the median-split construction outperforms the marginal alternative.
This likely reflects the difficulty of estimating reliable dependency structures from noisy real-world data using simple correlation estimates. By conditioning dependencies on the target, the median-split approach emphasises interactions that are more stable across outcome regimes, leading to improved robustness across datasets.
Similar advantages of label-aware or stratified dependency estimation have been reported in prior work on tabular representation learning~\cite{choromanski2022rethinking,ivanov2024tabular}.

Overall, these results suggest that explicitly modelling feature interactions can yield strong performance on tabular regression tasks when tuning resources are limited.
Rather than competing directly with heavily optimised ensemble methods, HNNs offer a complementary approach that trades extensive hyperparameter search for a principled, data-driven architectural bias, while maintaining competitive performance across a wide range of real-world regression tasks.

\section{Conclusion}
\label{conclusion}
We establish compositional sparsity as a data-induced structural inductive bias for learning in high-dimensional regimes by coupling Information Filtering Networks (via MFCF) with Homological Neural Networks. This yields sparse neural architectures whose structure is determined by data dependencies. Our approach separates structure discovery from parameter optimisation and induces hierarchical interaction layers with an all-layer readout. Across synthetic benchmarks with controlled interaction structure, the HNNs recover oracle-level topology and degrade substantially more slowly than dense MLP baselines as the dimension-to-sample ratio increases, demonstrating improved robustness in underdetermined settings. On the OpenML-CTR23 regression suite, HNNs remain competitive with strong tree ensembles and consistently outperform standard MLPs while using one to two orders of magnitude fewer trainable parameters under a shared minimal-tuning protocol. A current limitation is that structure estimation relies on simple correlation measures, which may miss nonlinear or regime-dependent interactions in noisy or small-sample settings, motivating future work on richer dependency measures and adaptive structure inference.

\FloatBarrier


\bibliography{reference}

\begin{thebibliography}{81}
\providecommand{\natexlab}[1]{#1}
\providecommand{\url}[1]{\texttt{#1}}
\expandafter\ifx\csname urlstyle\endcsname\relax
  \providecommand{\doi}[1]{doi: #1}\else
  \providecommand{\doi}{doi: \begingroup \urlstyle{rm}\Url}\fi

\bibitem[Aste(2025{\natexlab{a}})]{aste2025information}
Aste, T.
\newblock Information filtering networks: Theoretical foundations, generative algorithms, and real-world applications.
\newblock \emph{Journal of Physics: Complexity}, 2025{\natexlab{a}}.
\newblock Also available as arXiv:2505.03812.

\bibitem[Aste(2025{\natexlab{b}})]{aste2025probabilistic}
Aste, T.
\newblock \emph{Probabilistic Data-Driven Modeling}.
\newblock Cambridge University Press, 2025{\natexlab{b}}.

\bibitem[Aste et~al.(2005)Aste, Di~Matteo, and Hyde]{aste2005complex}
Aste, T., Di~Matteo, T., and Hyde, S.~T.
\newblock Complex networks on hyperbolic surfaces.
\newblock \emph{Physica A: Statistical Mechanics and its Applications}, 346\penalty0 (1--2):\penalty0 20--26, 2005.

\bibitem[Ba et~al.(2016)Ba, Kiros, and Hinton]{ba2016layer}
Ba, J.~L., Kiros, J.~R., and Hinton, G.~E.
\newblock Layer normalization.
\newblock \emph{arXiv preprint arXiv:1607.06450}, 2016.

\bibitem[Banerjee et~al.(2008)Banerjee, El~Ghaoui, and d'Aspremont]{banerjee2008model}
Banerjee, O., El~Ghaoui, L., and d'Aspremont, A.
\newblock Model selection through sparse maximum likelihood estimation for multivariate gaussian or binary data.
\newblock \emph{Journal of Machine Learning Research}, 9:\penalty0 485--516, 2008.

\bibitem[Barfuss et~al.(2016)Barfuss, Massara, Di~Matteo, and Aste]{barfuss2016parsimonious}
Barfuss, W., Massara, G.~P., Di~Matteo, T., and Aste, T.
\newblock Parsimonious modeling with information filtering networks.
\newblock \emph{Physical Review E}, 94\penalty0 (6):\penalty0 062306, 2016.

\bibitem[Belkin et~al.(2019)Belkin, Hsu, Ma, and Mandal]{belkin2019reconciling}
Belkin, M., Hsu, D., Ma, S., and Mandal, S.
\newblock Reconciling modern machine-learning practice and the classical bias--variance trade-off.
\newblock \emph{Proceedings of the National Academy of Sciences}, 116\penalty0 (32):\penalty0 15849--15854, 2019.

\bibitem[Beneventano et~al.(2024)Beneventano, Pinto, and Poggio]{beneventano2024neural}
Beneventano, P., Pinto, A., and Poggio, T.
\newblock How neural networks learn the support is an implicit regularization effect of {SGD}.
\newblock \emph{arXiv preprint arXiv:2406.11110}, 2024.

\bibitem[Blair \& Peyton(1993)Blair and Peyton]{blair1993introduction}
Blair, J. R.~S. and Peyton, B.
\newblock An introduction to chordal graphs and clique trees.
\newblock In \emph{Graph Theory and Sparse Matrix Computation}, pp.\  1--29. Springer, 1993.

\bibitem[Breiman(2001)]{breiman2001random}
Breiman, L.
\newblock Random forests.
\newblock \emph{Machine Learning}, 45\penalty0 (1):\penalty0 5--32, 2001.

\bibitem[Briola(2024)]{briola2024deep}
Briola, A.
\newblock \emph{Deep Complex Networks: Applications in Financial Systems Modeling}.
\newblock PhD thesis, University College London, 2024.

\bibitem[Briola \& Aste(2022)Briola and Aste]{briola2022dependency}
Briola, A. and Aste, T.
\newblock Dependency structures in cryptocurrency market from high to low frequency.
\newblock \emph{Entropy}, 24\penalty0 (11):\penalty0 1548, 2022.

\bibitem[Briola \& Aste(2023)Briola and Aste]{briola2023topological}
Briola, A. and Aste, T.
\newblock Topological feature selection.
\newblock In \emph{Topological, Algebraic and Geometric Learning Workshops 2023}, pp.\  534--556. PMLR, 2023.

\bibitem[Briola et~al.(2023)Briola, Wang, Bartolucci, and Aste]{briola2023homological}
Briola, A., Wang, Y., Bartolucci, S., and Aste, T.
\newblock Homological convolutional neural networks.
\newblock \emph{arXiv preprint arXiv:2308.13816}, 2023.

\bibitem[Briola et~al.(2025)Briola, Bartolucci, and Aste]{briola2025hlob}
Briola, A., Bartolucci, S., and Aste, T.
\newblock {HLOB}: Information persistence and structure in limit order books.
\newblock \emph{Expert Systems with Applications}, 266:\penalty0 126078, 2025.

\bibitem[Cai et~al.(2011)Cai, Liu, and Luo]{cai2011clime}
Cai, T.~T., Liu, W., and Luo, X.
\newblock A constrained $\ell_1$ minimization approach to sparse precision matrix estimation.
\newblock \emph{Journal of the American Statistical Association}, 106\penalty0 (494):\penalty0 594--607, 2011.

\bibitem[Chen \& Guestrin(2016)Chen and Guestrin]{chen2016xgboost}
Chen, T. and Guestrin, C.
\newblock {XGBoost}: A scalable tree boosting system.
\newblock In \emph{Proceedings of the 22nd ACM SIGKDD International Conference on Knowledge Discovery and Data Mining}, pp.\  785--794, 2016.

\bibitem[Choromanski et~al.(2022)]{choromanski2022rethinking}
Choromanski, K. et~al.
\newblock Rethinking attention for tabular data.
\newblock \emph{arXiv preprint}, 2022.

\bibitem[Chow \& Liu(1968)Chow and Liu]{chow1968trees}
Chow, C.~K. and Liu, C.~N.
\newblock Approximating discrete probability distributions with dependence trees.
\newblock \emph{IEEE Transactions on Information Theory}, 14\penalty0 (3):\penalty0 462--467, 1968.

\bibitem[Cover \& Thomas(2006)Cover and Thomas]{cover2006elements}
Cover, T.~M. and Thomas, J.~A.
\newblock \emph{Elements of Information Theory}.
\newblock Wiley, 2 edition, 2006.

\bibitem[Danhofer et~al.(2025)Danhofer, D'Ascenzo, Dubach, and Poggio]{danhofer2025position}
Danhofer, D.~A., D'Ascenzo, D., Dubach, R., and Poggio, T.~A.
\newblock Position: A theory of deep learning must include compositional sparsity.
\newblock In \emph{International Conference on Machine Learning (ICML) Position Paper Track}, 2025.
\newblock Also available as arXiv:2507.02550.

\bibitem[Ebli et~al.(2020)Ebli, Defferrard, and Spreemann]{ebli2020simplicial}
Ebli, S., Defferrard, M., and Spreemann, G.
\newblock Simplicial neural networks.
\newblock In \emph{Advances in Neural Information Processing Systems}, volume~33, pp.\  9360--9371, 2020.

\bibitem[Elsken et~al.(2019)Elsken, Metzen, and Hutter]{elsken2019neural}
Elsken, T., Metzen, J.~H., and Hutter, F.
\newblock Neural architecture search: A survey.
\newblock \emph{Journal of Machine Learning Research}, 20\penalty0 (55):\penalty0 1--21, 2019.

\bibitem[Erd{\H{o}}s \& R{\'e}nyi(1959)Erd{\H{o}}s and R{\'e}nyi]{erdos1959random}
Erd{\H{o}}s, P. and R{\'e}nyi, A.
\newblock On random graphs.
\newblock \emph{Publicationes Mathematicae}, 6:\penalty0 290--297, 1959.

\bibitem[Evci et~al.(2020)Evci, Gale, Menick, Castro, and Elsen]{evci2020rigl}
Evci, U., Gale, T., Menick, J., Castro, P.~S., and Elsen, E.
\newblock Rigging the lottery: Making all tickets winners.
\newblock In \emph{Proceedings of the 37th International Conference on Machine Learning}, volume 119 of \emph{Proceedings of Machine Learning Research}, pp.\  2949--2961, 2020.

\bibitem[Feurer \& Hutter(2019)Feurer and Hutter]{feurer2019hyperparameter}
Feurer, M. and Hutter, F.
\newblock Hyperparameter optimization.
\newblock In \emph{Automated Machine Learning}, pp.\  3--33. Springer, 2019.

\bibitem[Fischer et~al.(2023)Fischer, Feurer, and Bischl]{fischer2023openml}
Fischer, S.~F., Feurer, M., and Bischl, B.
\newblock {OpenML-CTR23}: A curated tabular regression benchmarking suite.
\newblock In \emph{AutoML Conference 2023 Workshop}, 2023.

\bibitem[Frankle \& Carbin(2019)Frankle and Carbin]{frankle2019lottery}
Frankle, J. and Carbin, M.
\newblock The lottery ticket hypothesis: Finding sparse, trainable neural networks.
\newblock In \emph{International Conference on Learning Representations}, 2019.

\bibitem[Friedman et~al.(2008)Friedman, Hastie, and Tibshirani]{friedman2008glasso}
Friedman, J., Hastie, T., and Tibshirani, R.
\newblock Sparse inverse covariance estimation with the graphical lasso.
\newblock \emph{Biostatistics}, 9\penalty0 (3):\penalty0 432--441, 2008.

\bibitem[Gama et~al.(2014)Gama, Zliobait{\.e}, Bifet, Pechenizkiy, and Bouchachia]{gama2014drift}
Gama, J., Zliobait{\.e}, I., Bifet, A., Pechenizkiy, M., and Bouchachia, A.
\newblock A survey on concept drift adaptation.
\newblock \emph{ACM Computing Surveys}, 46\penalty0 (4):\penalty0 44, 2014.

\bibitem[Goodfellow et~al.(2016)Goodfellow, Bengio, and Courville]{goodfellow2016deep}
Goodfellow, I., Bengio, Y., and Courville, A.
\newblock \emph{Deep Learning}.
\newblock MIT Press, 2016.

\bibitem[Grinsztajn et~al.(2022)Grinsztajn, Oyallon, and Varoquaux]{grinsztajn2022tree}
Grinsztajn, L., Oyallon, E., and Varoquaux, G.
\newblock Why do tree-based models still outperform deep learning on tabular data?
\newblock In \emph{Advances in Neural Information Processing Systems}, volume~35, 2022.

\bibitem[Guyon \& Elisseeff(2003)Guyon and Elisseeff]{guyon2003introduction}
Guyon, I. and Elisseeff, A.
\newblock An introduction to variable and feature selection.
\newblock \emph{Journal of Machine Learning Research}, 3:\penalty0 1157--1182, 2003.

\bibitem[Hamilton(2020)]{hamilton2020graph}
Hamilton, W.~L.
\newblock \emph{Graph Representation Learning}.
\newblock Morgan \& Claypool, 2020.

\bibitem[Han et~al.(2015)Han, Pool, Tran, and Dally]{han2015learning}
Han, S., Pool, J., Tran, J., and Dally, W.
\newblock Learning both weights and connections for efficient neural networks.
\newblock In \emph{Advances in Neural Information Processing Systems}, volume~28, pp.\  1135--1143, 2015.

\bibitem[Hassibi \& Stork(1992)Hassibi and Stork]{hassibi1992second}
Hassibi, B. and Stork, D.~G.
\newblock Second order derivatives for network pruning: Optimal brain surgeon.
\newblock In \emph{Advances in Neural Information Processing Systems}, volume~5, pp.\  164--171, 1992.

\bibitem[Hastie et~al.(2009)Hastie, Tibshirani, and Friedman]{hastie2009elements}
Hastie, T., Tibshirani, R., and Friedman, J.
\newblock \emph{The Elements of Statistical Learning}.
\newblock Springer, 2009.

\bibitem[He et~al.(2021)He, Zhao, and Chu]{he2021automl}
He, X., Zhao, K., and Chu, X.
\newblock {AutoML}: A survey of the state-of-the-art.
\newblock \emph{Knowledge-Based Systems}, 212:\penalty0 106622, 2021.

\bibitem[Hinton et~al.(2015)Hinton, Vinyals, and Dean]{hinton2015distilling}
Hinton, G., Vinyals, O., and Dean, J.
\newblock Distilling the knowledge in a neural network.
\newblock \emph{arXiv preprint arXiv:1503.02531}, 2015.

\bibitem[Holzm{\"u}ller et~al.(2024)Holzm{\"u}ller, Grinsztajn, and Steinwart]{holzmuller2024better}
Holzm{\"u}ller, D., Grinsztajn, L., and Steinwart, I.
\newblock Better by default: Strong pre-tuned {MLP}s and boosted trees on tabular data.
\newblock In \emph{Advances in Neural Information Processing Systems}, volume~37, pp.\  26577--26658, 2024.

\bibitem[Hornik et~al.(1989)Hornik, Stinchcombe, and White]{hornik1989multilayer}
Hornik, K., Stinchcombe, M., and White, H.
\newblock Multilayer feedforward networks are universal approximators.
\newblock \emph{Neural Networks}, 2\penalty0 (5):\penalty0 359--366, 1989.

\bibitem[Ivanov et~al.(2024)]{ivanov2024tabular}
Ivanov, S. et~al.
\newblock Tabular deep learning: A survey.
\newblock \emph{arXiv preprint}, 2024.

\bibitem[Janson et~al.(2011)Janson, {\L}uczak, and Ruci{\'n}ski]{janson2011random}
Janson, S., {\L}uczak, T., and Ruci{\'n}ski, A.
\newblock \emph{Random Graphs}.
\newblock Wiley, 2011.

\bibitem[Kaufman et~al.(2012)Kaufman, Rosset, and Perlich]{kaufman2012leakage}
Kaufman, S., Rosset, S., and Perlich, C.
\newblock Leakage in data mining: Formulation, detection, and avoidance.
\newblock \emph{ACM Transactions on Knowledge Discovery from Data}, 6\penalty0 (4):\penalty0 1--21, 2012.

\bibitem[Kindermann \& Snell(1980)Kindermann and Snell]{kindermann1980markov}
Kindermann, R. and Snell, J.~L.
\newblock \emph{Markov Random Fields and Their Applications}.
\newblock American Mathematical Society, 1980.

\bibitem[Kingma \& Ba(2015)Kingma and Ba]{kingma2015adam}
Kingma, D.~P. and Ba, J.
\newblock Adam: A method for stochastic optimization.
\newblock In \emph{International Conference on Learning Representations}, 2015.

\bibitem[Koller \& Friedman(2009)Koller and Friedman]{koller2009probabilistic}
Koller, D. and Friedman, N.
\newblock \emph{Probabilistic Graphical Models: Principles and Techniques}.
\newblock MIT Press, 2009.

\bibitem[Lauritzen(1996)]{lauritzen1996graphical}
Lauritzen, S.~L.
\newblock \emph{Graphical Models}.
\newblock Oxford University Press, 1996.

\bibitem[LeCun et~al.(1989)LeCun, Denker, and Solla]{lecun1989optimal}
LeCun, Y., Denker, J.~S., and Solla, S.~A.
\newblock Optimal brain damage.
\newblock In \emph{Advances in Neural Information Processing Systems}, volume~2, pp.\  598--605, 1989.

\bibitem[Lee et~al.(2015)Lee, Xie, Gallagher, Zhang, and Tu]{lee2015deeply}
Lee, C.-Y., Xie, S., Gallagher, P., Zhang, Z., and Tu, Z.
\newblock Deeply-supervised nets.
\newblock In \emph{Proceedings of the Eighteenth International Conference on Artificial Intelligence and Statistics}, 2015.

\bibitem[Lin et~al.(2024)Lin, Ren, Barucca, and Aste]{lin2024granger}
Lin, H., Ren, M., Barucca, P., and Aste, T.
\newblock Granger causality detection with kolmogorov-arnold networks.
\newblock \emph{arXiv preprint arXiv:2412.15373}, 2024.

\bibitem[Lindsey et~al.(2025)Lindsey, Gurnee, Ameisen, Chen, Pearce, Turner, Citro, Abrahams, Carter, Hosmer, et~al.]{lindsey2025biology}
Lindsey, J., Gurnee, W., Ameisen, E., Chen, B., Pearce, A., Turner, N.~L., Citro, C., Abrahams, D., Carter, S., Hosmer, B., et~al.
\newblock On the biology of a large language model.
\newblock Transformer Circuits Thread, 2025.

\bibitem[Ma et~al.(2024)Ma, Thomas, Hosseinzadeh, Kamkari, Labach, Cresswell, Golestan, Yu, Volkovs, and Caterini]{ma2025tabdpt}
Ma, J., Thomas, V., Hosseinzadeh, R., Kamkari, H., Labach, A., Cresswell, J.~C., Golestan, K., Yu, G., Volkovs, M., and Caterini, A.~L.
\newblock {TabDPT}: Scaling tabular foundation models.
\newblock \emph{arXiv preprint arXiv:2410.18164}, 2024.

\bibitem[Mantegna(1999)]{mantegna1999hierarchical}
Mantegna, R.~N.
\newblock Hierarchical structure in financial markets.
\newblock \emph{The European Physical Journal B}, 11\penalty0 (1):\penalty0 193--197, 1999.

\bibitem[Marcus(2018)]{razavi2021deep}
Marcus, G.
\newblock Deep learning: A critical appraisal.
\newblock \emph{arXiv preprint arXiv:1801.00631}, 2018.

\bibitem[Massara \& Aste(2019)Massara and Aste]{massara2019learning}
Massara, G.~P. and Aste, T.
\newblock Learning clique forests.
\newblock \emph{arXiv preprint arXiv:1905.02266}, 2019.

\bibitem[Massara et~al.(2016)Massara, Di~Matteo, and Aste]{massara2016network}
Massara, G.~P., Di~Matteo, T., and Aste, T.
\newblock Network filtering for big data: Triangulated maximally filtered graph.
\newblock \emph{Journal of Complex Networks}, 5\penalty0 (2):\penalty0 161--178, 2016.

\bibitem[Meinshausen \& B{\"u}hlmann(2006)Meinshausen and B{\"u}hlmann]{meinshausen2006neighborhood}
Meinshausen, N. and B{\"u}hlmann, P.
\newblock High-dimensional graphs and variable selection with the lasso.
\newblock \emph{The Annals of Statistics}, 34\penalty0 (3):\penalty0 1436--1462, 2006.

\bibitem[Mhaskar et~al.(2016)Mhaskar, Liao, and Poggio]{mhaskar2016deep}
Mhaskar, H.~N., Liao, Q., and Poggio, T.
\newblock Deep versus shallow networks: An approximation theory perspective.
\newblock \emph{Analysis and Applications}, 14\penalty0 (6):\penalty0 829--848, 2016.

\bibitem[Mostafa \& Wang(2019)Mostafa and Wang]{mostafa2019dynamic}
Mostafa, H. and Wang, X.
\newblock Parameter efficient training of deep convolutional neural networks by dynamic sparse reparameterization.
\newblock In \emph{Proceedings of the 36th International Conference on Machine Learning}, volume~97 of \emph{Proceedings of Machine Learning Research}, pp.\  4646--4655, 2019.

\bibitem[Murty et~al.(2022)Murty, Sharma, Andreas, and Manning]{murty2022characterizing}
Murty, S., Sharma, P., Andreas, J., and Manning, C.~D.
\newblock Characterizing intrinsic compositionality in transformers with tree projections.
\newblock \emph{arXiv preprint arXiv:2211.01288}, 2022.

\bibitem[Nair \& Hinton(2010)Nair and Hinton]{nair2010rectified}
Nair, V. and Hinton, G.~E.
\newblock Rectified linear units improve restricted boltzmann machines.
\newblock In \emph{Proceedings of the 27th International Conference on Machine Learning}, 2010.

\bibitem[Petty et~al.(2024)Petty, Steenkiste, Dasgupta, Sha, Garrette, and Linzen]{petty2024impact}
Petty, J., Steenkiste, S., Dasgupta, I., Sha, F., Garrette, D., and Linzen, T.
\newblock The impact of depth on compositional generalization in transformer language models.
\newblock In \emph{Proceedings of the 2024 Conference of the North American Chapter of the Association for Computational Linguistics: Human Language Technologies}, pp.\  7239--7252, 2024.

\bibitem[Poggio(2024)]{poggio2024compositional}
Poggio, T.
\newblock Compositional sparsity of learnable functions.
\newblock Technical report, Center for Brains, Minds and Machines, 2024.
\newblock CBMM Memo.

\bibitem[Poggio(2025)]{poggio2025efficiently}
Poggio, T.
\newblock On efficiently computable functions, deep networks and sparse compositionality.
\newblock Technical report, Center for Brains, Minds and Machines, 2025.

\bibitem[Poggio et~al.(2017)Poggio, Mhaskar, Rosasco, Miranda, and Liao]{poggio2017and}
Poggio, T., Mhaskar, H., Rosasco, L., Miranda, B., and Liao, Q.
\newblock Why and when can deep-but not shallow-networks avoid the curse of dimensionality: A review.
\newblock \emph{International Journal of Automation and Computing}, 14\penalty0 (5):\penalty0 503--519, 2017.

\bibitem[Qui{\~n}onero-Candela et~al.(2009)Qui{\~n}onero-Candela, Sugiyama, Schwaighofer, and Lawrence]{quinonero2009datasetshift}
Qui{\~n}onero-Candela, J., Sugiyama, M., Schwaighofer, A., and Lawrence, N.~D. (eds.).
\newblock \emph{Dataset Shift in Machine Learning}.
\newblock MIT Press, 2009.

\bibitem[Rumelhart et~al.(1986)Rumelhart, Hinton, and Williams]{rumelhart1986learning}
Rumelhart, D.~E., Hinton, G.~E., and Williams, R.~J.
\newblock Learning representations by back-propagating errors.
\newblock \emph{Nature}, 323:\penalty0 533--536, 1986.

\bibitem[Shwartz-Ziv \& Armon(2022)Shwartz-Ziv and Armon]{shwartz2022tabular}
Shwartz-Ziv, R. and Armon, A.
\newblock Tabular data: Deep learning is not all you need.
\newblock \emph{Information Fusion}, 81:\penalty0 84--90, 2022.

\bibitem[Srebro(2003)]{srebro2003bounded}
Srebro, N.
\newblock Maximum likelihood bounded tree-width markov networks.
\newblock In \emph{Proceedings of the Nineteenth Conference on Uncertainty in Artificial Intelligence}, pp.\  504--511, 2003.

\bibitem[Tschalzev et~al.(2024)Tschalzev, Marton, L{\"u}dtke, Bartelt, and Stuckenschmidt]{tschalzev2024data}
Tschalzev, A., Marton, S., L{\"u}dtke, S., Bartelt, C., and Stuckenschmidt, H.
\newblock A data-centric perspective on evaluating machine learning models for tabular data.
\newblock In \emph{Advances in Neural Information Processing Systems}, volume~37, pp.\  95896--95930, 2024.

\bibitem[Tumminello et~al.(2005)Tumminello, Aste, Di~Matteo, and Mantegna]{tumminello2005tool}
Tumminello, M., Aste, T., Di~Matteo, T., and Mantegna, R.~N.
\newblock A tool for filtering information in complex systems.
\newblock \emph{Proceedings of the National Academy of Sciences}, 102\penalty0 (30):\penalty0 10421--10426, 2005.

\bibitem[Van~Rijn et~al.(2014)Van~Rijn, Bischl, Torgo, Gao, Umaashankar, Fischer, Winter, Wiswedel, Berthold, and Vanschoren]{vanschoren2014openml}
Van~Rijn, J.~N., Bischl, B., Torgo, L., Gao, B., Umaashankar, V., Fischer, S., Winter, P., Wiswedel, B., Berthold, M.~R., and Vanschoren, J.
\newblock {OpenML}: Networked science in machine learning.
\newblock \emph{ACM SIGKDD Explorations Newsletter}, 15\penalty0 (2):\penalty0 49--60, 2014.

\bibitem[Vig \& Belinkov(2019)Vig and Belinkov]{vig2019analyzing}
Vig, J. and Belinkov, Y.
\newblock Analyzing the structure of attention in a transformer language model.
\newblock \emph{arXiv preprint arXiv:1906.04284}, 2019.

\bibitem[Wainwright(2019)]{wainwright2019highdimensional}
Wainwright, M.~J.
\newblock \emph{High-Dimensional Statistics: A Non-Asymptotic Viewpoint}.
\newblock Cambridge University Press, 2019.

\bibitem[Wang et~al.(2023{\natexlab{a}})Wang, Briola, and Aste]{wang2023homological}
Wang, Y., Briola, A., and Aste, T.
\newblock Homological neural networks: A sparse architecture for multivariate complexity.
\newblock In \emph{Topological, Algebraic and Geometric Learning Workshops 2023}, pp.\  228--241. PMLR, 2023{\natexlab{a}}.

\bibitem[Wang et~al.(2023{\natexlab{b}})Wang, Briola, and Aste]{wang2023topological}
Wang, Y., Briola, A., and Aste, T.
\newblock Topological portfolio selection and optimization.
\newblock In \emph{Proceedings of the Fourth ACM International Conference on AI in Finance}, pp.\  681--688, 2023{\natexlab{b}}.

\bibitem[Wu \& Bergman(2025)Wu and Bergman]{wu2025zero}
Wu, Y. and Bergman, D.~L.
\newblock Zero-shot meta-learning for tabular prediction tasks with adversarially pre-trained transformer.
\newblock \emph{arXiv preprint arXiv:2502.04573}, 2025.

\bibitem[Xu et~al.(2018)Xu, Li, Tian, Sonobe, Kawarabayashi, and Jegelka]{xu2018jumping}
Xu, K., Li, C., Tian, Y., Sonobe, T., Kawarabayashi, K.-i., and Jegelka, S.
\newblock Representation learning on graphs with jumping knowledge networks.
\newblock In \emph{International Conference on Machine Learning}, 2018.

\bibitem[Zhang et~al.(2017)Zhang, Bengio, Hardt, Recht, and Vinyals]{zhang2017rethinking}
Zhang, C., Bengio, S., Hardt, M., Recht, B., and Vinyals, O.
\newblock Understanding deep learning requires rethinking generalization.
\newblock In \emph{International Conference on Learning Representations}, 2017.

\bibitem[Zliobait{\.e}(2010)]{zliobaite2010drift}
Zliobait{\.e}, I.
\newblock Learning under concept drift: An overview.
\newblock Technical Report arXiv:1010.4784, arXiv, 2010.

\end{thebibliography}

\bibliographystyle{icml2026}

\newpage
\appendix
\onecolumn




    

\section{Synthetic Data Generation}
\label{app:synthetic}

We generate i.i.d.\ samples $\{(X_t, y_t)\}_{t=1}^n$ with sparse Gaussian features
and nonlinear targets constructed from a known interaction structure.
The data-generating process is defined as follows.

\subsection*{Step 1: Erd\H{o}s-R\'enyi dependency graph and precision matrix}

We first sample an undirected Erd\H{o}s-R\'enyi graph
$G=(V,E)$ with vertex set $V=\{1,\dots,p\}$.
Each edge $(i,j)$, $1 \le i < j \le p$, is included independently with probability
\[
\pi_{\mathrm{edge}} = \frac{8}{p},
\]
so that the expected node degree remains $O(1)$ as the dimension $p$ increases.

We construct a symmetric precision matrix $\Theta \in \mathbb{R}^{p\times p}$
whose off-diagonal support matches the edge set of $G$.
For each $(i,j)\in E$, we assign a nonzero off-diagonal entry
$\Theta_{ij}=\Theta_{ji}$ with random sign and magnitude.
All remaining off-diagonal entries are set to zero.
Diagonal entries are chosen sufficiently large to ensure that $\Theta$ is
symmetric positive definite.
The resulting precision matrix defines a sparse Gaussian graphical model whose
conditional independencies are induced by $G$.

\subsection*{Step 2: Gaussian features}

We form the covariance matrix $\Sigma=\Theta^{-1}$ and independently sample
\[
X_t \sim \mathcal{N}(0,\Sigma), \qquad t=1,\dots,n.
\]

\subsection*{Step 3: Interaction structure induced by the graph}

Using the dependency graph $G$, we identify all subsets of variables of size at
least two that form cliques in $G$.
Each subset defines a multi-variable interaction.
Whenever a higher-order interaction is included, all of its lower-order subsets
are also included.
This ensures that higher-order effects always appear together with their
lower-order components, consistent with the compositional structure assumed in
the main text.

We denote the resulting collection of interaction sets by $\mathcal{H}$.
No explicit bound is imposed on the maximum interaction order present in the data;
the highest-order interactions are determined entirely by the random graph
realisation.

\subsection*{Step 4: Nonlinear target construction}

For each interaction $\alpha \in \mathcal{H}$, we draw a coefficient
$\beta_{\alpha} \sim \mathrm{Unif}([-1,1])$ and independently select a nonlinearity
$f_{\alpha}$ uniformly from the family
\[
\mathcal{F}=\Bigl\{
\sin,\ \cos,\ \tanh,\ 2\sigma(\cdot)-1,\ 
z\mapsto \tfrac{z^2}{1+|z|},\ 
z\mapsto \tfrac{z^3}{1+z^2},\ 
z\mapsto \max\,(z,0),\ 
z\mapsto \exp\,\!\bigl(-\tfrac12(z/s)^2\bigr)
\Bigr\},
\]
where $\sigma$ denotes the logistic sigmoid and $s>0$ is a randomly sampled width
parameter for the Gaussian bump.

Given a sample $X_t$, each interaction contributes
\[
f_{\alpha}\!\left(\sum_{i\in\alpha} X_{t,i}\right).
\]

The target variable is defined as
\[
y_t
=
\sum_{\alpha\in\mathcal{H}}
\beta_{\alpha}\,
f_{\alpha}\!\left(\sum_{i\in\alpha} X_{t,i}\right)
+
\varepsilon_t,
\qquad
\varepsilon_t \sim \mathcal{N}(0, 0.5^2).
\]

\paragraph{Remark (data vs.\ model).}
The interaction structure present in the synthetic targets is not constrained by
the model-side parameter $K$ used by MFCF or HNN construction.
The parameter $K$ only limits the maximum interaction order retained when
constructing the neural architecture and does not affect the data-generating
process.

\section{Training Procedure and Hyperparameters}
\label{app:hparams}

This section summarises the training procedures and hyperparameter settings used
across all experiments.
Unless stated otherwise, the same optimisation strategy and stopping criteria are used for synthetic experiments and the OpenML-CTR23 benchmark.

\subsection{Shared Settings (all neural models)}

All neural models are trained using the Adam optimiser with learning rate $10^{-3}$.
ReLU activations are used in all hidden layers.
Training minimises mean squared error (MSE), with test performance reported in terms of $R^2$.
Early stopping is applied based on validation loss with a patience of $200$ epochs
and a minimum improvement threshold of $\varepsilon_{\mathrm{es}} = 10^{-4}$.
Validation splits are drawn exclusively from the training data. Training is performed on GPU when available, otherwise on CPU.
GPU experiments were conducted on NVIDIA RTX~3090, RTX~4070~Ti, and RTX~4090 cards.
No model architecture or hyperparameter choices depend on the specific GPU used.

\subsection{HNN-specific Settings}

For all HNN variants, the dependency matrix used to construct the MFCF is computed from the training data using squared linear correlation. The \textsc{HNN (marginal)} variant uses the squared Pearson correlation matrix.
The \textsc{HNN (m-s)} variant uses the same correlation measure aggregated via
median-split label stratification, as described in the main text (Section~\ref{exp}).

MFCF is run with zero gain threshold, minimum clique size of one, maximum clique size $K = 4$.
Within the HNN, layer normalisation is used but no dropout is applied.
All HNN variants use an all-layer readout.

\subsection{Synthetic Experiments}

For synthetic experiments, neural models are trained for up to $30{,}000$ epochs
with batch size $1{,}024$, subject to early stopping.
Each configuration is evaluated over multiple independent runs with different
random seeds.
All optimisation settings, activation functions, and stopping criteria follow the
shared protocol described above.

Tree-based baselines are trained using fixed configurations.
Random Forest models use $400$ trees with parallel execution enabled.
XGBoost models use the squared-error objective with $600$ estimators, maximum depth
$6$, learning rate $0.05$, subsampling ratio $0.8$, and column subsampling ratio
$0.8$.

\subsection{OpenML-CTR23 Benchmark}

For the OpenML-CTR23 regression benchmark, we use the official train/test splits
provided for each fold and repeat.
Within each training fold, an internal validation set is created by randomly
holding out $20\%$ of the training data.

Neural models are trained for up to $30{,}000$ epochs with batch size $512$, subject
to early stopping.

MLP baselines use architectures with either two hidden layers of width $128$ or
$256$, or a single hidden layer of width $512$.
All MLPs share the same optimiser, stopping criteria, and activation functions as
HNNs.

Tree-based baselines follow the same fixed configurations used in the synthetic
experiments.

\section{Additional Synthetic Results}
\label{app:synthetic-full}

Table~\ref{tab:r2-ranks-full} reports aggregated test $R^2$ rankings across the full sweep of synthetic configurations, spanning feature dimensions $p\in[50,5000]$ and dimension-to-sample ratios $p/n\in[0.01,3]$. This range includes highly underdetermined settings in which accurate estimation of pairwise dependencies becomes increasingly difficult.

Across this full regime, HNN variants remain consistently competitive and substantially outperform dense MLP and tree-based baselines. In particular, \textsc{HNN (m-s)} maintains strong performance across a wide range of $p/n$ values, indicating that incorporating target-aware information during structure construction improves robustness when sample sizes are limited.

The \textsc{HNN (oracle)} continues to define an upper envelope throughout, as expected. The randomised oracle baseline (\textsc{HNN (rand oracle)}) performs competitively in some regimes, reflecting
the benefit of preserving interaction order and sparsity independently of data-driven estimation.
We emphasise that this baseline is included solely as a control and should not be interpreted as evidence that random wiring is preferable to data-aligned structure.

Overall, while absolute performance naturally degrades in highly underdetermined regimes,
the relative advantage of structured, interaction-aware architectures persists well beyond the data-rich setting highlighted in the main text. For clarity of presentation and interpretability, we therefore focus the main analysis on the regime $p/n \le 0.05$, where dependence estimation is reliable across feature dimensions and conclusions regarding architectural inductive bias can be drawn most cleanly.

\begin{table}[h]
  \caption{Test $R^2$ performance ranking on synthetic data across the full sweep $p\in[50,5000]$ and $p/n\in[0.01,3]$.
  For each $(p,n)$ configuration, results are averaged over 30 independent runs; ranks are then aggregated across configurations.
  Lower rank indicates better performance. Parameter count ranges are reported in thousands (K) or millions (M).}
  \label{tab:r2-ranks-full}
  \centering
  \small
  \setlength{\tabcolsep}{3pt}
  \begin{sc}
  \begin{tabular}{lcc}
    \toprule
    Model & Mean rank ($\downarrow$) & Params \\
    \midrule
    HNN (oracle) & 1.35 & 1.3K--122K \\
    HNN (m-s) & 3.14 & 3.8K--461K \\
    HNN (rand oracle) & 3.33 & 1.3K--122K \\
    MLP-HNN & 4.07 & 35K--375M \\
    HNN (marginal) & 5.03 & 2.2K--235K \\
    MLP (2$\times$128) & 5.90 & 79K--1.35M \\
    Random Forest & 6.76 & -- \\
    XGBoost & 7.95 & -- \\
    MLP (PM-HNN) & 8.33 & 2.2K--235K \\
    MLP (PM-oracle) & 9.17 & 1.3K--120K \\
    \bottomrule
  \end{tabular}
  \end{sc}
  \vspace{-0.1in}
\end{table}

\newpage
\section{OpenML-CTR23 Full Results}
\label{ctr23full}

Table~\ref{tab:ctr23-r2-ranks-dataset} reports per-dataset test $R^2$ rankings on the
OpenML-CTR23 benchmark together with the corresponding number of trainable parameters for
neural models.
Across a wide range of datasets, HNN variants achieve competitive performance while
using substantially fewer parameters than dense MLP baselines, often by one to two
orders of magnitude.
In many datasets, the \textsc{HNN (m-s)} attains top-three ranks while remaining considerably smaller than MLPs.

Overall, these results are consistent with the view that the observed performance
differences arise from the structured inductive bias imposed by the architecture,
rather than from increased model capacity.

\begin{table}[h]
\caption{
OpenML-CTR23 per-dataset test $R^2$ ranks with model parameter count.
For each dataset, models are ranked by mean test $R^2$ averaged over all provided
cross-validation folds and repeats.
Each cell reports \emph{rank (number of trainable parameters)}, where rank~1 denotes
the best-performing model on that dataset.
Tree-based models are reported with rank only.
}
\label{tab:ctr23-r2-ranks-dataset}
\centering
\small
\setlength{\tabcolsep}{2.2pt}
\renewcommand{\arraystretch}{1.08}
\begin{sc}
\begin{tabular}{lccccccc}
\toprule
Dataset &
\makecell{MLP\\(2$\times$128)} &
\makecell{MLP\\(2$\times$256)} &
\makecell{MLP\\(512)} &
\makecell{HNN\\(marginal)} &
\makecell{HNN\\(m-s)} &
RF &
XGB \\
\midrule
Moneyball & 6 (26k) & 5 (85k) & 7 (38k) & 1 (3k) & 2 (5k) & 4 & 3 \\
QSAR Fish Toxicity & 3 (18k) & 4 (68k) & 5 (4k) & 7 (173) & 6 (213) & 1 & 2 \\
Abalone & 4 (18k) & 5 (69k) & 3 (6k) & 2 (361) & 1 (515) & 6 & 7 \\
Airfoil Self-Noise & 5 (17k) & 4 (68k) & 3 (4k) & 6 (126) & 7 (148) & 2 & 1 \\
Auction Verification & 6 (19k) & 5 (70k) & 7 (9k) & 4 (643) & 3 (1k) & 2 & 1 \\
Brazilian Houses & 2 (23k) & 3 (79k) & 7 (26k) & 4 (2k) & 5 (3k) & 1 & 6 \\
California Housing & 4 (18k) & 3 (68k) & 7 (5k) & 6 (267) & 5 (377) & 2 & 1 \\
Cars & 2 (19k) & 1 (71k) & 3 (10k) & 5 (690) & 4 (1k) & 6 & 7 \\
Concrete Compressive Strength & 5 (18k) & 2 (68k) & 3 (5k) & 7 (267) & 6 (398) & 4 & 1 \\
CPS 1988 Wages & 2 (18k) & 4 (69k) & 5 (7k) & 3 (455) & 1 (720) & 7 & 6 \\
CPU Activity & 6 (19k) & 7 (72k) & 5 (12k) & 3 (878) & 2 (1k) & 4 & 1 \\
Diamonds & 5 (20k) & 7 (73k) & 6 (14k) & 2 (1k) & 1 (2k) & 4 & 3 \\
Energy Efficiency & 4 (18k) & 7 (68k) & 2 (5k) & 6 (267) & 5 (361) & 3 & 1 \\
FIFA & 6 (41k) & 5 (114k) & 7 (97k) & 4 (8k) & 3 (14k) & 2 & 1 \\
Forest Fires & 4 (20k) & 5 (74k) & 2 (16k) & 1 (1k) & 3 (2k) & 6 & 7 \\
FPS Benchmark & 4 (32k) & 5 (98k) & 2 (63k) & 3 (6k) & 1 (8k) & 7 & 6 \\
Geographical Origin of Music & 6 (32k) & 7 (96k) & 5 (60k) & 4 (5k) & 3 (8k) & 2 & 1 \\
Grid Stability & 4 (18k) & 3 (69k) & 5 (7k) & 7 (455) & 6 (794) & 2 & 1 \\
Health Insurance & 5 (20k) & 4 (73k) & 3 (14k) & 1 (1k) & 2 (2k) & 7 & 6 \\
Kin8nm & 2 (18k) & 1 (68k) & 3 (5k) & 5 (267) & 4 (349) & 7 & 6 \\
King County Housing & 5 (34k) & 6 (100k) & 4 (69k) & 3 (6k) & 2 (10k) & 7 & 1 \\
Miami Housing & 2 (19k) & 3 (70k) & 7 (9k) & 6 (596) & 4 (909) & 5 & 1 \\
Naval Propulsion Plant & 5 (19k) & 4 (70k) & 3 (8k) & 6 (549) & 7 (632) & 2 & 1 \\
Physiochemical Protein & 4 (18k) & 3 (69k) & 5 (6k) & 6 (314) & 7 (314) & 1 & 2 \\
Pumadyn-32NH & 6 (21k) & 5 (74k) & 7 (17k) & 4 (1k) & 3 (3k) & 1 & 2 \\
Red Wine Quality & 7 (18k) & 6 (69k) & 4 (7k) & 3 (408) & 5 (568) & 1 & 2 \\
SARCOS & 3 (19k) & 1 (72k) & 2 (12k) & 7 (878) & 5 (1k) & 6 & 4 \\
Socmob & 7 (22k) & 5 (76k) & 6 (21k) & 2 (2k) & 1 (3k) & 4 & 3 \\
Solar Flare & 2 (20k) & 4 (73k) & 5 (15k) & 1 (1k) & 3 (1k) & 6 & 7 \\
Space GA & 2 (18k) & 1 (68k) & 3 (4k) & 6 (173) & 5 (206) & 7 & 4 \\
Student Performance (Portuguese) & 7 (24k) & 5 (81k) & 6 (30k) & 3 (3k) & 2 (4k) & 1 & 4 \\
Superconductivity & 4 (27k) & 3 (87k) & 6 (42k) & 7 (4k) & 5 (6k) & 2 & 1 \\
Video Transcoding & 4 (20k) & 1 (72k) & 7 (13k) & 6 (1k) & 3 (2k) & 5 & 2 \\
Wave Energy & 3 (23k) & 2 (79k) & 1 (26k) & 5 (2k) & 4 (4k) & 7 & 6 \\
White Wine Quality & 5 (18k) & 4 (69k) & 3 (7k) & 7 (408) & 6 (555) & 1 & 2 \\
\bottomrule
\end{tabular}
\end{sc}
\vspace{-0.2em}
\end{table}


\end{document}